\def\year{2022}\relax
\documentclass[letterpaper]{article} 
\usepackage{aaai22}  
\usepackage{times}  
\usepackage{helvet}  
\usepackage{courier}  
\usepackage[hyphens]{url}  
\usepackage{graphicx} 
\usepackage{natbib}  
\usepackage{caption} 
\DeclareCaptionStyle{ruled}{labelfont=normalfont,labelsep=colon,strut=off} 
\usepackage{algorithm}
\usepackage{algorithmic}

\usepackage{newfloat}
\usepackage{listings}
\floatstyle{ruled}
\newfloat{listing}{tb}{lst}{}
\floatname{listing}{Listing}

\usepackage[utf8]{inputenc} 
\usepackage[T1]{fontenc}    
\usepackage{hyperref}       
\usepackage{url}            
\usepackage{booktabs}       
\usepackage{amsfonts}       
\usepackage{nicefrac}       
\usepackage{microtype}      
\usepackage{amsmath}
\usepackage{amssymb}
\usepackage{dsfont}
\usepackage{xcolor}
\usepackage{caption}
\usepackage{subcaption}
\usepackage{float}
\newcommand{\E}{\mathbb{E}}

\newcount\Comments  
\newcommand{\kibitz}[2]{\ifnum\Comments=1{\textcolor{#1}{#2}}\fi}

\newtheorem{theorem}{Theorem}[section]
\newtheorem{lemma}[theorem]{Lemma}

\title{MultiplexNet: Towards Fully Satisfied Logical Constraints in Neural Networks}
\author {
    Nicholas Hoernle,\textsuperscript{\rm 1}
    Rafael Michael Karampatsis, \textsuperscript{\rm 1}
    Vaishak Belle \textsuperscript{\rm 1}, 
    Kobi Gal  \textsuperscript{\rm 1, \rm 2}
}
\affiliations {
    \textsuperscript{\rm 1} University of Edinburgh\\
    \textsuperscript{\rm 2} Ben-Gurion University\\
}

\usepackage{bibentry}

\begin{document}

\maketitle

\begin{abstract}
We propose a novel way to incorporate expert knowledge into the training of deep neural networks.
Many approaches encode domain constraints directly into the network architecture, requiring non-trivial or domain-specific engineering. 
In contrast, our approach, called MultiplexNet, represents domain knowledge as a logical formula in disjunctive normal form (DNF) which is easy to encode and to elicit from human experts. 
It introduces a Categorical latent variable that learns to choose which constraint term optimizes the error function of the network and it compiles the constraints directly into the output of existing learning algorithms.
We demonstrate the efficacy of this approach empirically on several classical deep learning tasks, such as density estimation and classification in both supervised and unsupervised settings where prior knowledge about the domains was expressed as logical constraints. 
Our results show that the MultiplexNet approach learned to approximate unknown distributions well, often requiring fewer data samples than the alternative approaches.
In some cases, MultiplexNet finds better solutions than the baselines; or solutions that could not be achieved with the alternative approaches.
Our contribution is in encoding domain knowledge in a way that facilitates inference that is shown to be both efficient and general; and critically, our approach guarantees 100\% constraint satisfaction in a network's output. 
\end{abstract}


\section{Introduction}
\label{sec:introduction}

An emerging theme in the development of deep learning is to provide expressive tools that allow domain experts to encode their prior knowledge into the training of neural networks. 
For example, in a manufacturing setting, we may wish to encode that an actuator for a robotic arm does not exceed some threshold (e.g., causing the arm to move at a hazardous speed).
Another example is a self-driving car, where a controller should be known to operate within a predefined set of constraints (e.g., the car should always stop completely at a stop street).
In such \emph{safety critical} domains, machine learning solutions must guarantee to operate within distinct boundaries that are specified by experts~\citep{amodei2016concrete}.

One possible solution is to encode the relevant domain knowledge directly into a network's architecture which may require non-trivial and/or domain-specific engineering  ~\citep{goodfellow2016deep}. 
An alternative approach is to express domain knowledge as logical constraints which can then be used to train neural networks~\citep{xu2017semantic,fischer2018dl2,allen2020probabilistic}.  
These approaches compile the constraints into the loss function of the training algorithm, by quantifying the extent to which the output of the network violates the constraints.
This is appealing as logical constraints are easy to elicit from people. 
However, the solution outputted by the network is designed to minimize the loss function --- which combines both data and constraints --- rather than to guarantee the satisfaction of the domain constraints. 
Thus, representing constraints in the loss function is not suitable for safety critical domains where 100\% constraint satisfaction is desirable.

Safety critical settings are not the only application for domain constraints.
Another common problem in the training of large networks is that of data inefficiency.
Deep models have shown unprecedented performance on a wide variety of tasks but these come at the cost of large data requirements.%
    \footnote{For instance OpenAI's GPT-3 \citep{brown2020language} was trained on about $500$ billion tokens and ImageNet-21k, used to train the ViT network \citep{dosovitskiy2020image}, consists of 14 million images.}
For tasks where domain knowledge exists, learning algorithms should also use this knowledge to structure a network's training to reduce the data burden that is placed on the learning process~\citep{fischer2018dl2}.

This paper directly addresses these challenges by providing a new way of representing domain constraints directly in the output layer of a network that guarantees constraint satisfaction. 
The proposed approach represents domain knowledge as a logical formula in disjunctive normal form (DNF).  
It augments the output layer of an existing neural network to include a separate transformation for each term in the DNF formula.
We introduce a latent Categorical variable that selects the best transformation that optimizes the loss function of the data.
In this way, we are able to represent arbitrarily complex domain constraints in an automated manner, and we are also able to guarantee that the output of the network satisfies the specified constraints. 

We  show the efficacy of this {MultiplexNet} approach in three distinct experiments.
First, we present a density estimation task on synthetic data. 
It is a common goal in machine learning to draw samples from a target distribution, and deep generative models have shown to be flexible and powerful tools for solving this problem.
We show that by including domain knowledge, a model can learn to approximate an unknown distribution on fewer samples, and the model will (by construction) only produce samples that satisfy the domain constraints.
This experiment speaks to both the data efficiency and the guaranteed constraint satisfaction desiderata. 
Second, we present an experiment on the popular MNIST data set~\citep{lecun2010mnist} which combines structured data with domain knowledge.
We structure the digits in a similar manner to the MNIST experiment from~\citet{manhaeve2018deepproblog}; however, we train the network in an entirely label-free manner~\citep{stewart2017label}.
In our third experiment, we apply our approach to the well known image classification task on the CIFAR100 data set~\citep{krizhevsky2009learning}. 
Images are clustered according to ``super classes'' (e.g., both \textit{maple tree} and \textit{oak tree} fall under the super class \textit{tree}).
We follow the example of \citet{fischer2018dl2} and show that by including the knowledge that images within a super class are related, we can increase the classification accuracy at the super class level.

The paper contributes a novel and general way to integrate domain knowledge in the form of a logical specification into the training of neural networks.
We show that domain knowledge may be used to restrict the network's operating domain such that any output is guaranteed to satisfy the constraints; and in certain cases, the domain knowledge can help to train the networks on fewer samples of data.

\subsection{Problem Specification}
We consider a data set of $N$ i.i.d. samples from a hybrid (some mixture of discrete and/or continuous variables) probability density~\citep{belle2015probabilistic}.
Moreover, we assume that: (1) the data set was generated by some random process $p^*(x)$; and (2) there exists domain or expert knowledge, in the form of a logical formula $\Phi$, about the random process $p^*(x)$ that can express the domain where $p^*(x)$ is feasible (non-zero).
Both of these assumptions are summarised in Eq.~\ref{eq:domain_constraints}. 
In Eq.~\ref{eq:domain_constraints}, the notation $x \models \Phi$, denotes that the sample $x$ satisfies the formula $\Phi$~\citep{barrett2009handbook}. 
For example, if $\Phi := x > 3.5 \land y > 0$, and given some sample $(x, y) = (5, 2)$, we denote: $(x, y) \models \Phi$.

\begin{equation}
    \label{eq:domain_constraints}
    x \sim p^*(x) \implies x \models \Phi
\end{equation}

Our aim is to approximate $p^*(x)$ with some parametric model $p_\theta(x)$ and to incorporate the domain knowledge $\Phi$ into the maximum likelihood estimation of $\theta$, on the available data set.

Given knowledge of the constraints $\Phi$, we are interested in ways of integrating these constraints into the training of a network that approximates $p^*(x)$.
We desire an algorithm that does not require novel engineering to solve a reparameterisation of the network and moreover, especially salient for safety-critical domains, any sample $x$ from the model, $x \sim p_\theta(x)$, should imply that the constraints are satisfied.
This is an especially important aspect to consider when comparing this method to alternative approaches, namely \citet{fischer2018dl2} and \citet{xu2017semantic}, that do not give this same guarantee.

\section{Related Work}

The integration of domain knowledge into the training of neural networks is an emerging area of focus. 
Many previous studies attempt to translate logical constraints into a numerical loss. 
The two most relevant works in this line are the DL2 framework by~\citet{fischer2018dl2} and the Semantic Loss approach by~\citet{xu2017semantic}.
DL2 uses a loss term that trades off data with the domain knowledge. 
It defines a non-negative loss by interpreting the logical constraints using fuzzy logic and defining a measure that quantifies how far a network's output is from the nearest satisfying solution. 
Semantic Loss also defines a term that is added to the standard network loss.
Their loss function uses weighted model counting~\citep{chavira2008probabilistic} to evaluate the probability that a sample from a network's output satisfies some Boolean constraint formulation. 
We differ from both of these approaches in that we do not add a loss term to the network's loss function, rather we compile the constraints directly into its output. 
Furthermore, in contrast to the works above, any network output from MultiplexNet will satisfy the domain constraints, which is crucial in safety critical domains. 

It is also important to compare the expressiveness of the MultiplexNet constraints to those permitted by \citet{fischer2018dl2} and \citet{xu2017semantic}. 
In MultiplexNet, the constraints can consist of any quantifier-free linear arithmetic formula over the rationals.
Thus, variables can be combined over $+$ and $\geq$, and formulae over $\neg$, $\lor$ and $\land$. 
For example, $(x + y \geq 5) \land \neg (z \geq 5)$ but also $(x + y \geq z) \land (z > 5 \lor z < 3)$ are well defined formulae and therefore well defined constraints in our framework.
The expressiveness is significant --- for example, \citet{xu2017semantic} only allow for Boolean variables over $\{\neg, \land, \lor \}$.
While \citet{fischer2018dl2} allow non-Boolean variables to be combined over $\{ \geq, \leq \}$ and formulae to be used over $\{ \neg, \lor, \land \}$, it is not a probabilistic framework, but  one that is based on fuzzy logic. 
Thus, our work is probabilistic like the Semantic Loss~\citep{xu2017semantic}, but it is more expressive in that it also allows real-valued variables over summations too. 

\citet{hu2016harnessing} introduce ``iterative rule knowledge distillation'' which uses a student and teacher framework to balance constraint satisfaction on first order logic formulae with predictive accuracy on a classification task. 
During training, the student is used to form a constrained teacher by projecting its weights onto a subspace that ensures satisfaction of the logic.
The student is then trained to imitate the teacher's restricted predictions. 
\citet{hu2016harnessing} use soft logic~\citep{bach2017Hinge} to encode the logic, thereby allowing gradient estimation; however, the approach is unable to express rules that constrain real-valued outputs. 
Xsat~\citep{Fu2016XSat} focuses on the Satisfiability Modulo Theory (SMT) problem, which is concerned with deciding whether a (usually a quantifier-free form) formula in first-order logic is satisfied against a background arithmetic theory; similar to what we consider.
They present a means for solving SMT formulae but this is not differentiable.
\citet{manhaeve2018deepproblog} present a compelling method for integrating logical constraints, in the form of a ProbLog program, into the training of a network.
However, the networks are embedded into the logic (represented by a Sentential Decision Diagram~\citep{darwiche2011sdd}), as ``neural predicates'' and thus it is not clear how to handle the real-valued arithmetic constraints that we represent in MultiplexNet.

We also relate to work  on program synthesis \citep{lezama2009sketching,jha2010oracle,yu2017component,osera2019constraint} where the goal is to produce a valid program for a given set of constraints.
Here, the output of a program is designed to meet a given specification. 
These works differ from this paper as they don’t focus on the core problem of aiding training with the constraints and ensuring that the constraints are fully satisfied.

Other recent works have also explored how human expert knowledge can be used to guide a network's training. 
\citet{ross2018improving,ross2017right} explore how the robustness of an image classifier can be improved by regularizing input gradients towards regions of the image that contain information (as identified by a human expert).
They highlight the difficulty in eliciting expert knowledge from people but their technique is similar to the other works presented here in that the knowledge loss is still represented as an additive term to the standard network loss.
\citet{takeishi2020Knowledge} present an example of how the knowledge of relations of objects can be used to regularise a generative model. 
Again, the solution involves appending terms to the loss function, but they demonstrate that relational information can aid a learning algorithm.
Alternative works have also explored means for constraining the latent variables in a latent variable model~\cite{ganchev2010posterior,gracca2007expectation}. 
In contrast to this, we focus on constraining the output space of a generative model, rather than the latent space.

Finally, we mention work on the post-hoc verification of networks. 
Examples include the works of \citet{katz2017reluplex} and \citet{bunel2017unified} who present methods for validating whether a network will operate within the bounds of pre-defined restrictions. 
Our own work focuses on how to guarantee that the networks operate within given constraints, rather than how to validate their output.

\section{Incorporating Domain Constraints into Model Design}

We begin by describing how a satisfiability problem can be hard coded into the output of a network. 
We then present how any specification of knowledge can be compiled into a form that permits this encoding.
An overview of the proposed architecture with a general algorithm that details how to incorporate domain constraints into training a network can be found in Appendix B in the supplementary material.

\subsection{Satisfiability as Reparameterisation}
\label{sec:sat_as_reparam}

Let $\tilde{x}$ denote the unconstrained output of a network. 
Let $g$ be a network activation that is element-wise non-negative (for example an exponential function, or a ReLU~\citep{nair2010rectified} or Softplus~\citep{dugas2001incorporating} layer).
If the property to be encoded is a simple inequality $\Phi: \forall x \; c x \geq b$, it is sufficient to constrain $\tilde{x}$ to be non-negative by applying $g$ and thereafter applying a linear transformation $f$ such that:
$\forall \tilde{x}: c f(g(\tilde{x})) \geq b$.
In this case, ${f}$ can implement the transformation $f(z) = sgn(c)z + \frac{b}{c}$ where $sgn$ is the operator that returns the sign of $c$.
By construction we have:

\begin{equation}
    \label{eq:constrain_one}
        f(g(\tilde{x})) \models \Phi
\end{equation}

It follows that more complex conjunctions of constraints can be encoded by composing transformations of the form presented in Eq.~\ref{eq:constrain_one}.
We present below a few examples to demonstrate how this can be achieved for a number of common constraints (where $\tilde{x}$ always refers to the unconstrained output of the network):

\begin{align}
    \label{eq:common_operations_line1}
    a < x < b \; &\rightarrow \; x = -g(-g(\tilde{x}) + k(a, b)) + b \\
    \label{eq:common_operations_line2}
    x = c \; &\rightarrow \; x = c \\
    \label{eq:common_operations_line3}
    x_2 > h(x_1) \; &\rightarrow \; x_1 = \tilde{x_1} \; ; \; x_2 = h(x_1) + g(\tilde{x_2})
\end{align}

In Eq~\ref{eq:common_operations_line1}, we introduce the function $k(a, b)$. 
This is merely a function to compute the correct offset for a given activation $g$.
In the case of the Softplus function, which is the function used in all of our experiments, $k(a,b) = log(exp(b-a) -1)$.

In Section: Experiments, we implement three varied experiments that demonstrate how complex constraints can be constructed from this basic primitive in Eq.~\ref{eq:constrain_one}.
Conceptually, appending additional conjunctions to $\Phi$ serves to restrict the space that the output can represent.
However, in many situations domain knowledge will consist of complicated formulae that exist well beyond mere conjunctions of inequalities.

While conjunctions serve to restrict the space permitted by the network's output, disjunctions serve to increase the permissible space. 
For two terms $\phi_1$ and $\phi_2$ in $\phi_1 \lor \phi_2$ there exist three possibilities: namely, 
that $x \models \phi_1$ or $x \models \phi_2$ or $(x \models \phi_1) \land (x \models \phi_2)$.
Given the fact that any unconstrained network output can be transformed to satisfy some term $\phi_k$, we propose to introduce multiple transformations of a network's unconstrained output, each to model the different terms $\phi_k$.
In this sense, the network's output layer can be viewed as a multiplexor in a logical circuit that permits for a branching of logic.
If $h_1(\tilde{x})$ represents the transformation of $\tilde{x}$ that satisfies $\phi_1$ and $h_2(\tilde{x}) \models \phi_2$ then we know the output must also satisfy $\phi_1 \lor \phi_2$ by choosing either $h_1$ or $h_2$.
It is this branching technique for dealing with disjunctions that gives rise to the name of the approach: MultiplexNet.

We finally turn to the desideratum of allowing any Boolean formula over linear inequalities as the input for the domain constraints.
The suggested approach can represent conjunctions of constraints and disjunctions between these conjunctive terms, which is exactly a DNF representation.
Thus, the approach can be used with any transformed version of $\Phi$ that is in DNF~\citep{darwiche2002knowledge}.
We propose to use an off-the-shelf solver, e.g., Z3~\citep{de2008z3}, to provide the logical input to the algorithm that is in DNF.
We thus assume the domain knowledge $\Phi$ is expressed as:
\begin{equation}
    \label{eq:dnf_representation}
    \Phi = \phi_1 \lor \phi_2 \lor \hdots \lor \phi_k
\end{equation}

If $h_k$ is the branch of MultiplexNet that ensures the output of the network $x \models \phi_k$ then it follows by construction that $h_k(\tilde{x})\models \Phi$ for all $k \in [1, \dots, K]$.
For example, consider a network with a single real-valued output $\tilde{x} \in \mathbb{R}$. 
If the knowledge $\Phi := (x \geq 2) \lor (x \leq -2)$, we would then have the two terms $h_1(\tilde{x}) = g(\tilde{x}) + 2$ and $h_2(\tilde{x}) = -g(-\tilde{x}) - 2$. 
Here, $g$ is the network activation that is element-wise non-negative that was referred to in Section: Satisfiability as Reparameterisation.
It is clear that both $x_1 = h_1(\tilde{x})$ and $x_2 = h_2(\tilde{x})$ satisfy the formula $\Phi$.

\begin{lemma}
Suppose $\Phi$ is a quantifier free first-order formula in DNF over $\{x_1,\dots,x_J\}$ consisting of terms $\phi_1~\lor~\dots~\lor~\phi_K$.
Since each branch of MultiplexNet ($h_k$) is constructed to satisfy a specific term ($\phi_k$), by construction, the output of MultiplexNet will satisfy $\Phi$: $\{\hat{x}_1,\dots,\hat{x}_J\} \models \Phi$.
\end{lemma}

\subsection{MultiplexNet as a Latent Variable Problem}

MultiplexNet introduces a latent Categorical variable $k$ that selects among the different terms $\phi_k, k \in [1, \dots, K]$.
The model then incorporates a constraint transformation term $h_k$ conditional on the value of the Categorical variable. 

\begin{equation}
    \label{eq:introduce_latent_categorical_variable}
    p_\theta(x) = p_\theta(h_k(x) | k)p(k)
\end{equation}

A lower bound on the likelihood of the data can be obtained by introducing a variational approximation to the latent Categorical variable $k$.
This standard form of the variational lower bound (ELBO) is presented in Eq.~\ref{eq:elbo}.
\begin{equation}
    \begin{split}
        \label{eq:elbo}
        \log p_\theta (x) &\geq  \E_{q(k)}[\log p_\theta(h_k(x) | k) + \log p(k) - \log q(k)] \\
        &:= ELBO(x)
    \end{split}
\end{equation}

Gradient based methods require calculating the derivative of Eq.~\ref{eq:elbo}.
However, as $q(k)$ is a Categorical distribution, the standard reparameterisation trick cannot be applied~\citep{kingma2013auto}.
One possibility for dealing with this expectation is to use the score function estimator, as in REINFORCE~\citep{williams1992simple}; however, while the resulting estimator is unbiased, it has a high variance~\citep{mnih2014neural}.
It is also possible to replace the Categorical variable with a continuous approximation as is done by \citet{maddison2016concrete} and \citet{jang2016categorical}; or, if the dimensionality of the Categorical variable is small, it can be marginalised out as in~\cite{kingma2014semi}.
In the experiments in Section: Experiments, we follow \citet{kingma2014semi} and marginalise this variable,%
    \footnote{Although we note that the alternatives should also be explored.}
leading to the following learning objective:

\begin{multline}
    \label{eq:learning_objective}
    \mathcal{L}(\theta; x) = -\sum\limits_{k=1}^{K} q(k)\big[ \log p_\theta(h_k(x) | k) \\ + \log p(k) - \log q(k) \big]
\end{multline}

We show in Section: Experiments that this approach can be applied equally successfully for a generative modeling task (where the goal is density estimation) as for a discriminative task (where the goal is structured classification).
This helps to demonstrate the universal applicability of incorporating domain knowledge into the training of networks.

\section{Experiments}
\label{sec:experiments}

We apply MultiplexNet to three separate experimental domains. 
The first domain demonstrates a density estimation task on synthetic data when the number of available data samples are limited.
We show how the value of the domain constraints improves the training when the number of data samples decreases; this demonstrates the power of adding domain knowledge into the training pipeline.
The second domain applies MultiplexNet to labeling MNIST images in an unsupervised manner by exploiting a structured problem and data set.
We use a similar experimental setup to the MNIST experiment from DeepProbLog \citep{manhaeve2018deepproblog}; however, we present a natural integration with a generative model that is not possible with DeepProbLog.
The third experiment uses hierarchical domain knowledge to facilitate an image classification task taken from \citet{fischer2018dl2} and \citet{xu2017semantic}.
We show how the use of this knowledge can help to improve classification accuracy at the super class level.

\subsection{Synthetic Data}
\label{sec:illustrative_experiment}

\begin{figure}
  \centering
  \includegraphics[width=.39\textwidth]{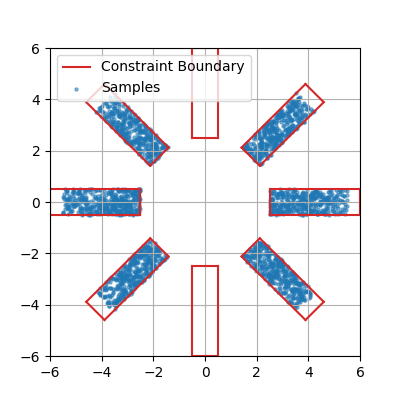}
  \caption{Simulated data from an unknown density. We assume that we know some constraints about the domain; these are represented by the red boxes. We aim to represent the unknown density, subject to the knowledge that the constraints must be satisfied.}
  \label{fig:data_samples_from_generated_data}
\end{figure}

\begin{figure*}
     \centering
     \begin{subfigure}[b]{.8\textwidth}
         \centering
         \includegraphics[width=\textwidth]{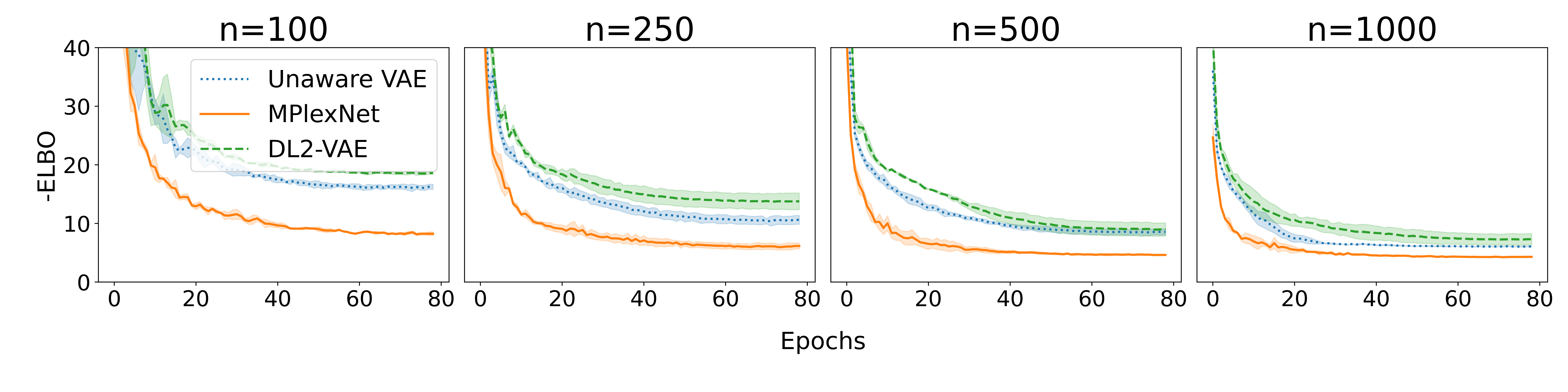}
         \caption{}
         \label{fig:synthetic_experiment_computational_results_subplot-a}
     \end{subfigure}
     
     \begin{subfigure}[b]{.8\textwidth}
         \centering
         \includegraphics[width=\textwidth]{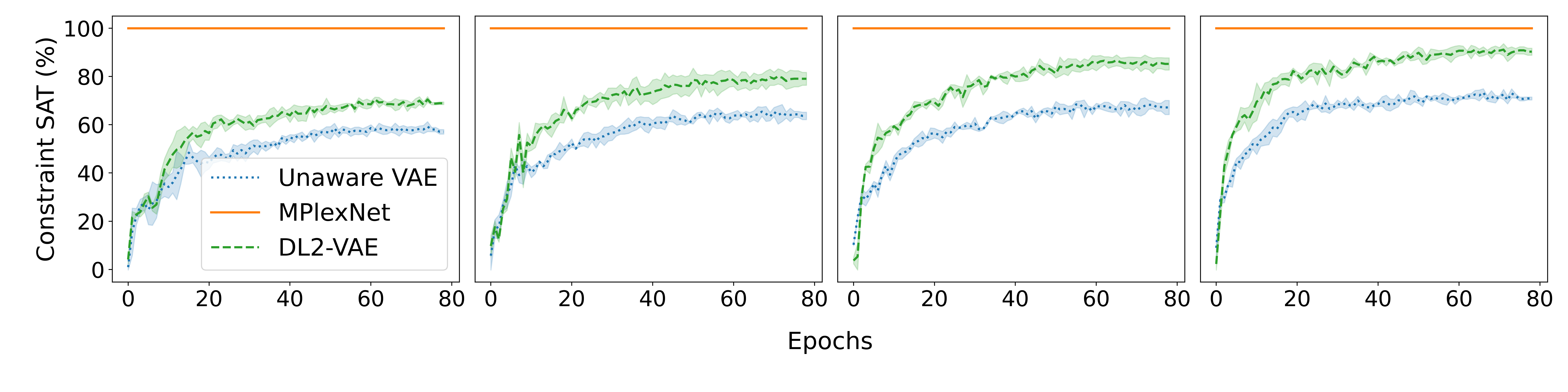}
         \caption{}
         \label{fig:synthetic_experiment_computational_results_subplot-b}
     \end{subfigure}

        \caption{Results from the synthetic data experiment (a) Negative lower bound to the held out likelihood of data (-ELBO). The MultiplexNet approach learns to represent the data with a higher likelihood, and faster than the baselines. (b) $\%$ of reconstruction samples from the VAE that obey the domain constraints. The MultiplexNet approach, by construction, can only generate samples within the specified constraints.}
        \label{fig:synthetic_experiment_computational_results}
\end{figure*}

In this illustrative experiment, we consider a target data set that consists of the six rectangular modes that are shown in Figure~\ref{fig:data_samples_from_generated_data}.
The samples from the true target density are shown, along with $8$ rectangular boxes in red.
The rectangular boxes represent the domain constraints for this experiment.
Here, we show that an expert might know where the data can exist but that the domain knowledge does not capture all of the details of the target density.
Thus, the network is still tasked with learning the intricacies of the data that the domain constraints fail to address (e.g., not all of the area within the constraints contains data).
However, we desire that the knowledge leads the network towards a better solution, and also to achieve this on fewer data samples from the true distribution.

This experiment represents a density estimation task and thus we use a likelihood-based generative model to represent the unknown target density, using both data samples and domain knowledge.
We use a variational autoencoder (VAE) which optimizes a lower bound to the marginal log-likelihood of the data. 
However, a different generative model, for example a normalizing flow~\citep{papamakarios2019normalizing} or a GAN~\citep{goodfellow2014generative}, could as easily be used in this framework.
We optimize Eq.~\ref{eq:learning_objective} where, for this experiment, the likelihood term $\log p_\theta ( \cdot \mid k)$ is replaced by the standard VAE loss.
Additional experimental details, as well as the full loss function, can be found in Appendix A.

We vary the size of the training data set with $N \in \{ 100, 250, 500, 1000 \}$ as the four experimental conditions.
We compare the lower bound to the marginal log-likelihood under three conditions: the MultiplexNet approach, as well as two baselines.
The first baseline (Unaware VAE) is a vanilla VAE that is unaware of the domain constraints. 
This scenario represents the standard setting where domain knowledge is simply ignored in the training of a deep generative network.
The second baseline (DL2-VAE) represents a method that appends a loss term to the standard VAE loss.
It is important to note that this approach, from DL2~\citep{fischer2018dl2}, does not guarantee that the constraints are satisfied (clearly seen in Figure~\ref{fig:synthetic_experiment_computational_results_subplot-b}).

Figure~\ref{fig:synthetic_experiment_computational_results} presents the results where we run the experiment on the specified range of training data set sizes.
The top plot shows the variational loss as a function of the number of epochs.
For all sizes of training data, the MultiplexNet loss on a test set can be seen to outperform the baselines. 
By including domain knowledge, we can reach a better result, and on fewer samples of data, than by not including the constraints. 
More important than the likelihood on held-out data is that the samples from the models' posterior should conform with the constraints.
Figure~\ref{fig:synthetic_experiment_computational_results_subplot-b} shows that the baselines struggle to learn the structure of the constraints.
While the MultiplexNet solution is unsurprising, the constraints are followed by construction, the comparison to the baselines is stark.
We also present samples from both the prior and the posterior for all of these models in Appendix A. 
In all of these, MultiplexNet learns to approximate the unknown density within the predefined boundaries of the provided constraints.
     
\subsection{MNIST - Label-free Structured Learning}
\label{sec:mnist_experiment}
We  demonstrate how a structured data set, in combination with the relevant domain knowledge, can be used to make novel inferences in that domain.
Here, we use a similar experiment to that from \citet{kingma2014semi} where we model the MNIST digit data set in an unsupervised manner.
Moreover, we take inspiration from \citet{manhaeve2018deepproblog} for constructing a structured data set were the images represent the terms in a summation (e.g., $image(2) + image(3) = 5$). 
However, we add to the complexity of the task by (1) using no labels for any of the images;%
    \footnote{In the MNIST experiment from \citet{manhaeve2018deepproblog}, the authors use the result of the summation as labels for the algorithm. We have no such analogy in this experiment and thus cannot use their DeepProbLog implementation as a baseline.}
and, (2) considering a generative task.

\citet{kingma2014semi} propose a generative model that reasons about the cluster assignment of a data point (a single image).
In particular, in their popular ``Model 2,'' they describe a generative model for an image $x$ such that the probability of the image pixel values are conditioned on a latent variable ($z$) and a class label ($y$): $p_\theta(x \mid z, y)p(z \mid y)p(y)$.
We can interpret this model using the MultiplexNet framework where the cluster assignment label $y = k$ implies that the image $x$ was generated from cluster $k$.
Given a reconstruction loss for image $x$, conditioned on class label $y$ ($\mathcal{L}(x, y)$), the domain knowledge in this setting is: $\Phi := \bigvee_{k=1}^{10} \mathcal{L}(x, y) \land (y = k)$.
We can successfully model the clustering of the data using this setup but there is no means for determining which label corresponds to which cluster assignment. 

We therefore propose to augment the data set such that each input is a quintuple of four images $(x_1, x_2, x_3, x_4)$ in the form $label(x_1) + label(x_2) = (label(x_3), label(x_4))$.
Here, the inputs $label(x_1)$ and $label(x_2)$ can be any integer from $0$ to $9$ and the result $(label(x_3), label(x_4))$ is a two digit number from $(00)$ to $(18)$.
While we do not know explicitly any of the cluster labels, we do know that the data conform to this standard.
Thus for all $i,j,k$ where $k=i + j$, the domain knowledge is of the form:

\begin{equation}
    \begin{split}
    \label{eq:mnist_domain_knowledge}
    \Phi &:= \bigvee_{i,j,k} \Big[ ( y_1 = i ) \land ( y_2 = j ) \land ( y_3 = \mathds{1}_{k>9} ) \\
         & \land ( y_4 = k \bmod 10 ) \bigwedge\limits_{n=1}^4 \mathcal{L}(x_n, y_n) \Big]
    \end{split}
\end{equation}

In this setting, the categorical variable in the MultiplexNet chooses among the $100$ combinations that satisfy $label(x_1) + label(x_2) = (label(x_3), label(x_4))$.
This experiment has similarities to DeepProbLog~\citep{manhaeve2018deepproblog} as the primitive $\mathcal{L}(x, y)$ is repeated for each digit.
In this sense, it is similar to the ``neural predicate'' used by \citet{manhaeve2018deepproblog}, and the MultiplexNet output layer implements what would be the logical program from DeepProbLog.
However, it is not clear how to implement this label-free, generative task within the DeepProbLog framework.

\begin{figure}
  \centering
  \includegraphics[width=.5\textwidth]{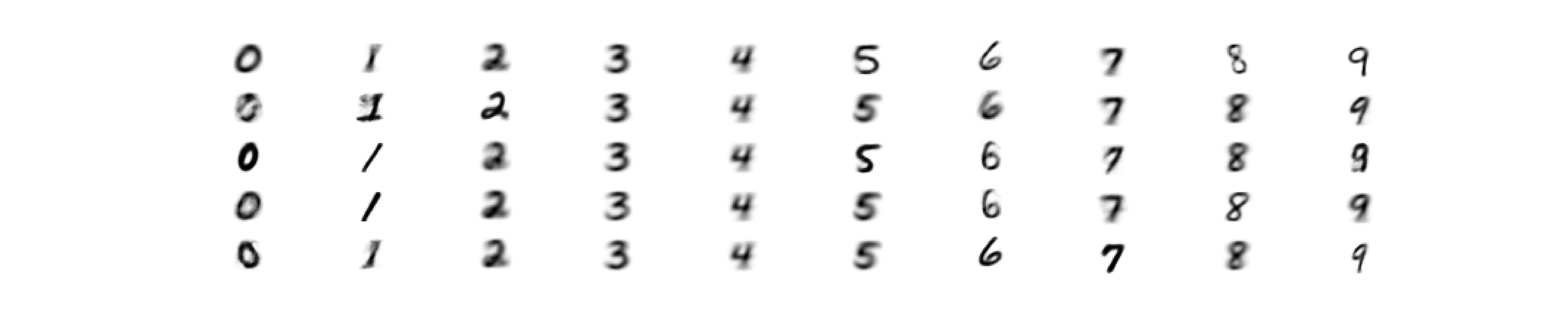}
  \caption{Reconstructed/Decoded samples from the prior, $z$, of the trained model where each column conditions on a different value for $y$. It can be seen that the model has learnt to represent all of the digits $[0-9]$  with the correct class label, even though no labels were supplied to the training process.}
  \label{fig:mnist_prior_samples}
\end{figure}

In Figure~\ref{fig:mnist_prior_samples}, we present samples from the prior, conditioned on the different class labels. 
The model is able to learn a class-conditional representation for the data, \textit{given no labels for the images}.
This is in contrast to a vanilla model (from \citet{kingma2014semi}) which does not use the structure of the data set to make inferences about the class labels.
We present these baseline samples as well as the experimental details and additional notes in Appendix A.
Empirically, the results from this experiment were sensitive to the network's initialisation and thus we report the accuracy of the top $5$ runs.
We selected the runs based on the loss (the ELBO) on a validation set 
(i.e., the labels were still not used in selecting the run). 
The accuracy of the inferred labels on held out data is $97.5 \pm 0.3$.

\subsection{Hierarchical Domain Knowledge on CIFAR100}

\begin{table*}
  \caption{Accuracy on class label prediction and super-class label prediction, and constraint satisfaction on CIFAR100 data set}
  \label{table:cifar100}
  \centering
  \begin{tabular}{l|ccc}
    \toprule
    Model     & Class Accuracy     & Super-class Accuracy & Constraint Satisfaction\\
    \midrule
    Vanilla ResNet          & $75.0 \pm (0.1)$ & $84.0 \pm (0.2)$ & $83.8 \pm (0.1)$ \\
    Vanilla ResNet (SC only)& NA               & $83.2 \pm (0.2)$ & NA \\
    Hierarchical Model      & $71.2 \pm (0.2)$ & $84.7 \pm (0.1)$ & $\mathbf{100.0 \pm (0.0)}$ \\
    DL2                     & $\mathbf{75.3 \pm (0.1)}$ & $84.3 \pm (0.1)$ & $85.8  \pm (0.2)$\\
    MultiplexNet            & $74.4 \pm (0.2)$ & $\mathbf{85.4 \pm (0.3)}$ & $\mathbf{100.0 \pm (0.0)}$         \\
    \bottomrule
  \end{tabular}
\end{table*}

The final experiment demonstrates how to encode hierarchical domain knowledge into the output layer of a network.
The CIFAR100 \citep{krizhevsky2009learning} data set consists of $100$ classes of images where the $100$ classes are in turn broken into $20$ super-classes (SC).
We wish to encode the belief that images from the same SC are semantically related.
Following the encoding in \citet{fischer2018dl2}, we consider constraints which specify that groups of classes should together be very likely or very unlikely.
For example, suppose that the SC label is \textit{trees} and the class label is \textit{maple}.
Our domain knowledge should state that the \textit{trees} group must be very likely even if there is uncertainty in the specific label \textit{maple}.
Intuitively, it is egregious to misclassify this example as a \textit{tractor} but it would be acceptable to make the mistake of \textit{oak}.
This can be implemented by training a network to predict first the SC for an unknown image and thereafter the class label, conditioned on the value for the SC.

We chose rather to implement this same knowledge using the MultiplexNet framework.
Let $x_{k} \in SC_i$ denote the output of a network that predicts the $k^{th}$ class label within the $i^{th}$ SC.
Let $\alpha \in [0, 1]$ denote the minimum requirement for a SC prediction (e.g., if  $\alpha = 0.95$, we require that a SC be predicted with probability $0.95$ or more).
The domain knowledge is:

\begin{equation}
    \label{eq:cifar100_domain_knowledge}
    \bigvee_{i=1}^{20} \left[ \bigwedge_{k \in SC_i} \Big( x_{k} > \log(\frac{\alpha}{1-\alpha}) + \log \sum\limits_{j \notin SC_i} \exp \{ x_j \} \Big) \right]
\end{equation}

Eq.~\ref{eq:cifar100_domain_knowledge} states that for all labels within a SC group, the unnormalised logits of the network should be greater than the normalised sum of the other labels belonging to the other SCs with a margin of $\log(\frac{\alpha}{1-\alpha})$.
We explain Eq.~\ref{eq:cifar100_domain_knowledge} further and present other experimental details in Appendix A.
This constraint places a semantic grouping on the data as the network is forced into a low entropy prediction at the super class level.

We compare the performance of MultiplexNet to three baselines and report the prediction accuracy on the fine class label as well that on the super class label. 
We use a Wide ResNet 28-10~\citep{zagoruyko2016wide} model in all of the experimental conditions. 
The first two baselines (Vanilla) only use the Wide ResNet model and are trained to predict the fine class and the super class labels respectively.
The second baseline (Hierarchical) is trained to predict the super class label and thereafter the fine class label, conditioned on the value for the super class label. 
This represents the bespoke engineering solution to this hierarchical problem.
The final baseline (DL2) implements the same logical specification that is used for MultiplexNet but uses the DL2 framework to append to the standard cross-entropy loss function.

Table~\ref{table:cifar100} presents the results for this experiment.
Firstly, it is important to note the difficulty of this task. 
The Vanilla ResNet that predicts only the super-class labels for the images under performs the baseline that is tasked with predicting the true class label.
Moreover, while the hierarchical baseline does outperform the vanilla models on the task of super-class prediction, this comes at a cost to the true class accuracy.
As the hierarchical baseline represents the bespoke engineering solution to the problem, it also achieves $100\%$ constraint satisfaction, but this comes at the cost of domain specific and custom implementation.
The MultiplexNet approach provides a slight improvement at the SC classification accuracy and importantly, the domain constraints are always met.
As the domain knowledge prioritizes the accuracy at the SC level, we note that the MultiplexNet approach does not outperform the Vanilla ResNet at the class accuracy.
Surprisingly, the DL2 baseline improves upon the class accuracy but it has a limited impact on the super class accuracy and on the constraint satisfaction.


\section{Limitations and Discussion}
\label{sec:limitations}
The limitations of the suggested approach relate to the technical specification of the domain knowledge and to the practical implementation of this knowledge.
We discuss first these two aspects and then we discuss a potential negative societal impact.

First, we require that experts be able to express precisely, in the form of a logical formula, the constraints that are valid for their domain.
This may not always be possible.
For example, given an image classification task, we may wish to describe our knowledge about the content of the images.
Consider an example where images contain pictures of \textit{dogs} and \textit{fish} and that we wish to express the knowledge that dogs must have four legs and fish must be in water.
It is not clear how these conceptual constraints would then be mapped to a pixel level for actual specification.
Moreover, it is entirely plausible to have images of dogs that do not include their legs, or images of fish where the fish is out of the water.
The logical statement itself is brittle in these instances and would serve to hinder the training, rather than to help it.
This example serves to present the inherent difficulty that is present when actually expressing robust domain knowledge in the form of logical formulae.

The second major limitation of this approach deals with the DNF requirement on the input formula. 
We require that knowledge be expressed in this form such that the ``or'' condition is controlled by the latent Categorical variable of MultiplexNet. 
It is well known that certain formulae have worst case representations in DNF that are exponential in the number of variables.
This is clearly undesirable in that the network would have to learn to choose among the exponentially many terms.

One of the overarching motivations for this work is to constrain networks for safety critical domains.
While constrained operation might be desired on many accounts, there may exist edge cases where an autonomously acting agent should act in an undesirable manner to avoid an even more undesirable outcome (a thought experiment of this spirit is the well known Trolley Problem~\citep{hammond2021learning}).
By guaranteeing that the operating conditions of a system be restricted to some range, our approach does encounter vulnerability with respect to  edge, and unforeseen, cases.
However, to counter this point,  we argue it is still necessary for experts to define the boundaries over the operation domain of a system in order to explicitly test and design for known worst case scenario settings.  


\section{Conclusions and Future Work}

This work studied  how logical knowledge in an expressive language could be used to constrain the output of a network.
It provides a new and general way to encode domain knowledge as logical constraints directly in the output layer of a network. 
Compared to  alternative approaches, we go beyond propositional logic by allowing for arithmetic operators in our constraints.
We are able to guarantee that the network output is 100\% compliant with the  domain constraints, which the alternative approaches, which append a ``constraint loss,'' are unable to match. Thus our approach is especially relevant for safety critical settings in which the network  must guarantee to operate within predefined constraints.
In a series of experiments we demonstrated that our approach leads to better results in terms of data efficiency (the amount of training data that is required for good performance), reducing the data burden that is placed on the training process.
In the future, we are excited about exploring the prospects for using this framework on downstream tasks, such as robustness to adversarial attacks.

\bibliography{bibliography}

\clearpage
\newpage

\appendix

\section{Appendix A: Additional Experimental Details}
\label{appendix:additional_experiment_details}

All code and data for repeating the experiments can be found at the repository at: \url{https://github.com/NickHoernle/semantic_loss}.
The image experiments (on MNIST and CIFAR100) were run on Nvidia GeForce RTX 2080 Ti devices and the synthetic data experiment was run on a 2015 MacBook Pro with processor specs: 2.2 GHz Quad-Core Intel Core i7.
The MNIST data set is available under the \textit{Creative Commons Attribution-Share Alike 3.0 license}; and CIFAR100 is available under the \textit{Creative Commons Attribution-Share Alike 4.0 license}.
The DL2 framework (available under an \textit{MIT License}), used in the baselines, is available from \url{https://github.com/eth-sri/dl2}.

In all experiments, the data were split into a train, validation and test set where the test set was held constant across the experimental conditions (e.g., in the CIFAR100 experiment, the same test set was used to compare MultiplexNet vs the vanilla models vs the DL2 model). 
In cases where model selection was performed (early stopping on  CIFAR100 and selection of the best runs from the MNIST experiment, we used the validation set to choose the best runs and/or models).
In these cases the validation set was extracted from the training data set prior to training (with $10\%$ of the data used for validation).
The standard test sets, given by MNIST and CIFAR100 were used for those experiments and an additional test set was generated for the synthetic experiment.

\subsection{Synthetic Data}
\label{appendix:synthetic_experiment_details}

\textbf{Deriving the Loss}

We first present the full derivation of the loss function that was used for this experiment.
We used a variational autoencoder (VAE) with a standard isotropic Gaussian prior.
The standard VAE loss is presented in Eq.~\ref{eq:standard_vae_loss}. 
In the below formulation, $x_i$ is a datapoint, $L$ is a minibatch size, and $z_i$ is a sample from the approximate posterior $q$.

\begin{equation}
    \label{eq:standard_vae_loss}
    \mathcal{L}(\theta) = -\sum\limits_{i=1}^L \log p(x_i \mid z_i) + \log p(z_i) - \log q(z_i \mid x_i) 
\end{equation}

We use an isotropic Gaussian likelihood for $\log p(x_i \mid z_i)$ and an isotropic Gaussian for the posterior.
Standard derivations (see \citet{kingma2013auto} for more details) allow the loss to be expressed as in Eq.\ref{eq:standard_vae_loss_simplified}.
In this equation, $f_{\theta}$ is the decoder model and it predicts the mean of the likelihood term.
A tunable parameter $\sigma$ controls the precision of the reconstructions -- this parameter was held constant for all experimental conditions.
The posterior distribution is a Gaussian with parameters $\sigma_q^2$ and $\mu_q$ that are output from the encoding network $q_\theta(x_i)$.

\begin{multline}
    \label{eq:standard_vae_loss_simplified}
    \mathcal{L}(\theta) = -\sum\limits_{i=1}^L \mathcal{N}(x_i; f_{\theta}(z_i), \sigma^2) + 0.5 * (1 + \log(\sigma_q^2)\\ - \mu_q^2 - \sigma_q^2)
\end{multline}

Finally, we present how the MultiplexNet loss uses $\mathcal{L}(\theta)$ in the transformation of the output layer of the network.
MultiplexNet takes as input the unconstrained network output $f_{\theta}(z_i)$ and it outputs the transformed (constrained) terms $h_k$ (for $K$ terms in the DNF constraint formulation) and the probability of each term $\pi_k$.
Let $\mathcal{L}_{k}(\theta)$ denote the same loss $\mathcal{L}(\theta)$ from Eq.~\ref{eq:standard_vae_loss_simplified} but with the raw output of the network $f_{\theta}(z_i)$ constrained by the constraint transformation $h_k$ (i.e., the likelihood term becomes: $\mathcal{N}(x_i; h_k(f_{\theta}(z_i)), \sigma^2)$)
The final loss then is presented in Eq.~\ref{eq:multiplexnetloss}.

\begin{equation}
    \label{eq:multiplexnetloss}
    MPlexNet(\theta) = \sum_{k=1}^{K} \pi_{k}  \left( \mathcal{L}_{h_k}(\theta) + \log \pi_{h_k} \right)
\end{equation}

\textbf{Samples from the Posterior}

\begin{figure*}
     \centering
     \begin{subfigure}[b]{\textwidth}
         \centering
         \includegraphics[width=\textwidth]{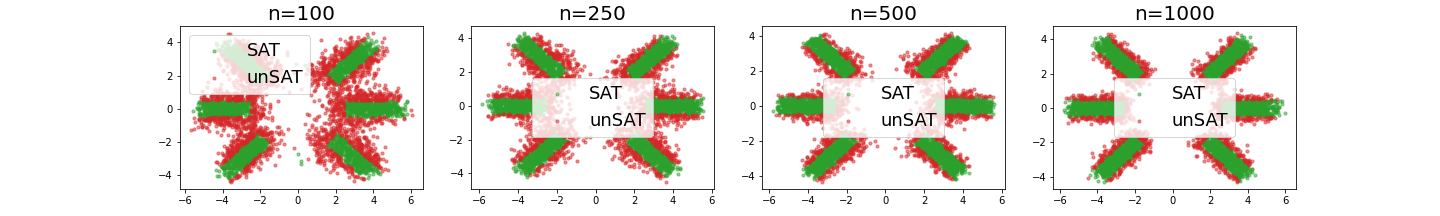}
         \caption{}
         \label{fig:synthetic_experiment_posterior_subplot-a}
     \end{subfigure}
     
     \begin{subfigure}[b]{\textwidth}
         \centering
         \includegraphics[width=\textwidth]{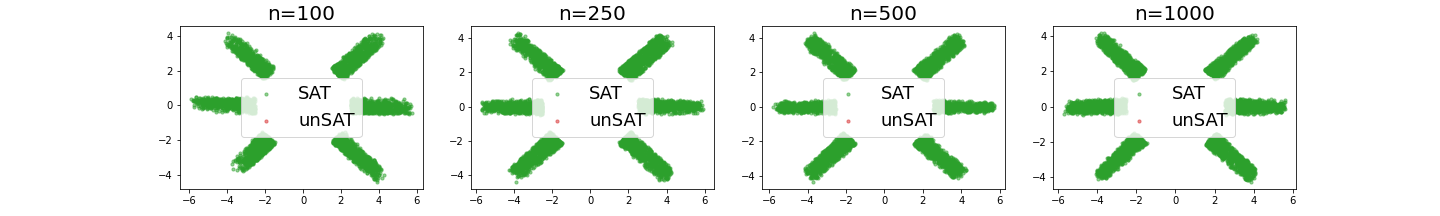}
         \caption{}
         \label{fig:synthetic_experiment_posterior_subplot-b}
     \end{subfigure}
    \caption{(a) Samples from the vanilla VAE posterior for different sizes of training data sets (b) Samples from the MultiplexNet VAE posterior for different sizes of training data sets.}
    \label{fig:synthetic_experiment_posterior}
\end{figure*}

Fig.~\ref{fig:synthetic_experiment_posterior} shows samples from the posterior of the two models -- these are the attempts of the VAE to reconstruct the input data.
It can clearly be seen that while MultiplexNet strictly adheres to the constraints, the baseline VAE approach fails to capture the constraint boundaries.

\textbf{Samples from the Prior}

\begin{figure*}
     \centering
     \begin{subfigure}[b]{\textwidth}
         \centering
         \includegraphics[width=\textwidth]{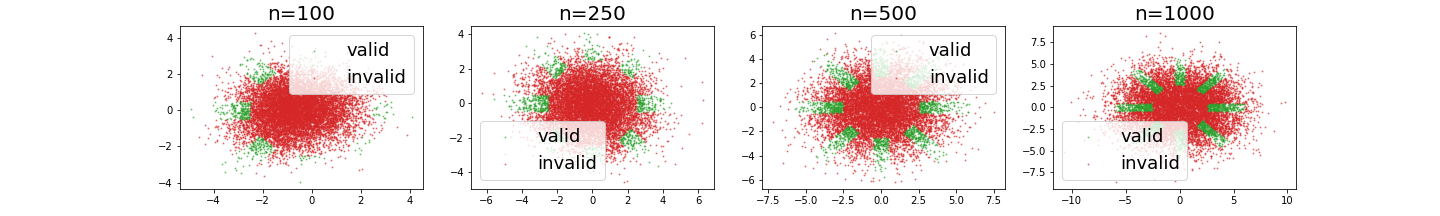}
         \caption{}
         \label{fig:synthetic_experiment_prior_subplot-a}
     \end{subfigure}
     
     \begin{subfigure}[b]{\textwidth}
         \centering
         \includegraphics[width=\textwidth]{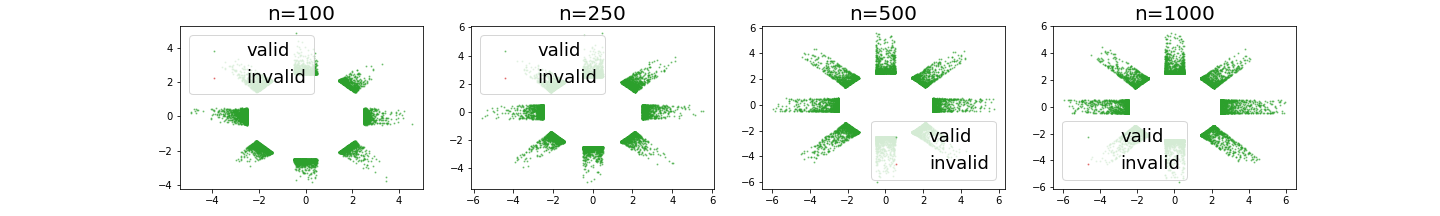}
         \caption{}
         \label{fig:synthetic_experiment_prior_subplot-b}
     \end{subfigure}

    \caption{(a) Samples from the vanilla VAE prior for different sizes of training data sets (b) Samples from the MultiplexNet VAE prior for different sizes of training data sets.}
    \label{fig:synthetic_experiment_prior}
\end{figure*}

Samples from the prior (Fig.~\ref{fig:synthetic_experiment_prior}) show how well a generative model has learnt the data manifold that it attempts to represent. 
We show these to demonstrate that in this case, the vanilla VAE fails to capture many of the complexities in the data distribution.
To sample from the prior for MultiplexNet, we randomly sample from the latent Categorical variable from MultiplexNet.
Hence, the two vertical modes (that contain no data in reality) have samples here.
This can easily be solved by introducing a trainable prior parameter over the Categorical variable as well -- an easy extension that we do not implement in this work.

\textbf{Network Architecture}

The default network used in these experiments was a feed-forward network with a single hidden layer for both the decoder and the encoder models. 
The dimensionality of the latent random variable was 15 and the hidden layer contained $50$ units.
ReLU activations were used unless otherwise stated.

\subsection{MNIST - Label-free Structured Learning}
\label{appendix:mnist_details}

\textbf{Deriving the Loss}

We follow the specification from \citet{kingma2014semi} where the likelihood of a single image, $x$, conditioned on a cluster assignment label $y$, is shown in Eq.~\ref{eq:mnist_kingma_loss}. 
Again, $z$ is a latent parameter, again assumed to follow an isotropic Gaussian distribution.

\begin{multline}
    \label{eq:mnist_kingma_loss}
    \log p(x, y) \geq \E_{q(z \mid x)} \big[\log p_{\theta}(x \mid y, z) \\ + \log p(z) - \log q(z \mid x)\big]
\end{multline}

We refer to the right hand side of Eq.~\ref{eq:mnist_kingma_loss} as $-V(x, y)$. 
Eq.~\ref{eq:mnist_kingma_loss} assumes knowledge of the label $y$, but this is unknown for our domain.
However, we can implement the knowledge from Eq.~\ref{eq:mnist_domain_knowledge} (in the main text) that specifies all $100$ possibilities for the image inference task.
Below we assume the data is of the form $image_i + image_j = (image_{k_1}, image_{k_2})$ where $label_i \in [0, \dots ,9]$, $label_j \in [0, \dots, 9]$, $label_{k_1} \in [0, 1]$ and $label_{k_2} \in [0, \dots, 9]$.
Finally, we let $\pi_h$ refer to the MultiplexNet Categorical selection variable that chooses which of the $100$ possible terms for $(i, j, k_1, k2)$ are present.
Following the MultiplexNet framework, the loss is then presented in Eq~\ref{eq:mnist_mplex_loss}:

\begin{multline}
    \label{eq:mnist_mplex_loss}
    \mathcal{L}(\theta) = \sum\limits_{i,j,k} \pi_h \big[ V(x_1, y_1 = i) + V(x_2, y_2 = j) + \\ V(x_3, y_3 = k_1) + V(x_4, y_4 = k_2) + \log \pi_h \big]
\end{multline}

\textbf{Samples from the Vanilla VAE prior from \citet{kingma2014semi}}

We repeat the mnist experiment with a vanilla VAE model (``Model 2'' from \citet{kingma2014semi}). 
Here we simply show that the model can capture the label clustering of the data but that it cannot, unsurprisingly, infer the class labels correctly from the data without considering the fact that the data set has been structured:

\begin{figure*}[H]
  \centering
  \includegraphics[width=.9\textwidth]{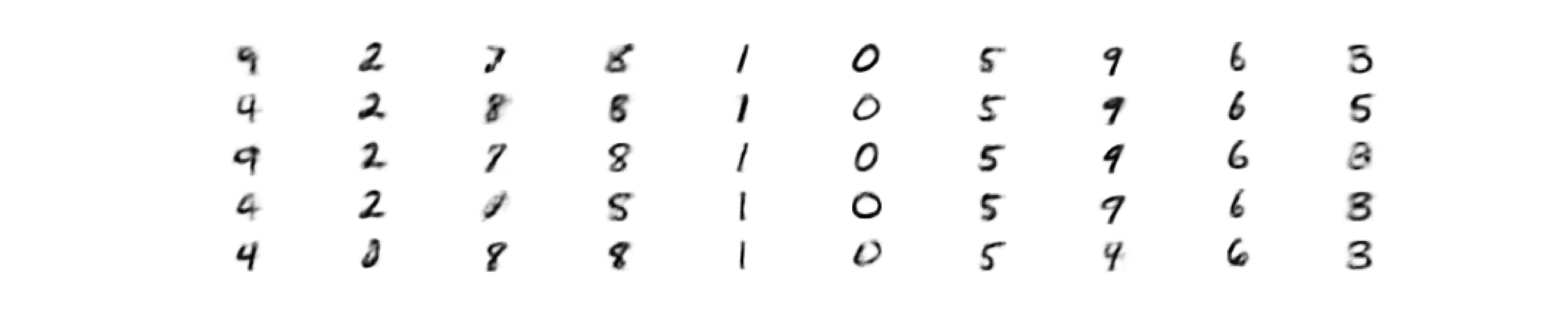}
  \caption{Reconstructed/Decoded samples from the prior, $z$, of ``Model 2'' from \citet{kingma2014semi}. The clustering of the data is clear but the model is unable to infer the correct class labels without considering the structured data set and domain knowledge.}
  \label{fig:mnist_kimna_prior_samples}
\end{figure*}

\textbf{Network Architecture}

The default network used in these experiments was a feed-forward network with two hidden layers for both the decoder and the encoder models.
The first hidden layer contained $250$ units and the second $100$ units. 
The dimensionality of the latent random variable was $50$.
ReLU activations were used unless otherwise stated.

\subsection{Hierarchical Domain Knowledge on CIFAR100}
\label{appendix:cifar100_details}

\textbf{Deriving the Loss}

Following the encoding in \cite{fischer2018dl2}, we consider constraints which specify that groups of classes should together be very likely or very unlikely.
For example, suppose that the SC label is \textit{trees} and the class label is \textit{maple}.
Our domain knowledge should state that the \textit{trees} group must be very likely even if there is uncertainty in the specific label \textit{maple}.
\citet{fischer2018dl2} use the following logic to encode this belief (for the $20$ super classes that are present in CIFAR100):

\begin{multline}
    \label{eq:fisher_logic}
    (p_{people} < \epsilon \lor p_{people} > 1-\epsilon) \land \dots \land \\ (p_{trees} < \epsilon \lor p_{trees} > 1-\epsilon)
\end{multline}

This encoding is exactly the same as that presented in Eq~\ref{eq:fisher_logic_multiplex}.
However, we re-write this encoding in DNF such that it is compatible with MultiplexNet.

\begin{multline}
    \label{eq:fisher_logic_multiplex}
    (p_{people} > 1-\epsilon \land p_{trees} < \epsilon \land \dots) \lor \\ (p_{people} < \epsilon \land p_{trees} >  1-\epsilon \land \dots) \lor ... 
\end{multline}

A simplification on the above, as the classes and thus the super classes lie on a simplex, is the specification that $(p_{people} > 1-\epsilon)$ necessarily implies that $(p_{trees} < \epsilon \land \dots)$ holds too.
Thus the logic can again be simplified to:

\begin{multline}
    \label{eq:fisher_logic_multiplex_simplified}
    (p_{people} > 1-\epsilon) \lor ( p_{trees} >  1-\epsilon  \\ \lor ( p_{fish} >  1-\epsilon ) \lor ... 
\end{multline}

Again, as the probability values here lie on a simplex, we can represent a single constraint as in Eq~\ref{eq:expand_person_class}.
Here, $Z$ is the normalizing constant that ensures the final output values are a valid probability distribution over the class labels (computed with a softmax layer in practice). 
$Z = e^{baby} + e^{boy} + \dots + e^{cattle} + e^{tractor}$ for all $100$ class labels in CIFAR100.

\begin{multline}
    \label{eq:expand_person_class}
    (p_{people} > 1-\epsilon) \implies \frac{e^{baby}}{Z} +\frac{e^{boy}}{Z} + \frac{e^{girl}}{Z} \\ +\frac{e^{man}}{Z} + \frac{e^{woman}}{Z} > 1 - \epsilon
\end{multline}

Finally, we can simplify the right hand side of Eq~\ref{eq:expand_person_class} to obtain the following specification (for the $people$ super class, but the other SCs all follow via symmetry):

\begin{multline}
    \label{eq:constraint1}
     e^{baby} + e^{boy} + e^{girl} + e^{man} + e^{woman} >  \\
     \frac{1-\epsilon}{\epsilon} \left[ e^{beaver} + e^{couch} + \dots + e^{streetcar} \right]
\end{multline}

Note that the right hand side of Eq~\ref{eq:constraint1} contains the classes for all the other super classes but not including $people$. 
It thus contains $95$ labels in this example.

Studying each of the terms in $j \in [baby, boy, girl, man, woman]$ separately, and noting that $e^y$ is strictly positive, we obtain Eq~\ref{eq:constraint2}.
We use $y_j$ to denote a class label in the target super class (in this case $people$) and $y_i$ to refer to all other labels in all other super classes (SC).

\begin{equation}
    \label{eq:constraint2}
     e^{y_j} > \frac{1-\epsilon}{\epsilon} \left[ \sum\limits_{i \notin SC_{people}} e^{y_i} \right]
\end{equation}

As we are interested in constraining the unnormalized output of the network ($y_j$) in MultiplexNet, it is clear that we can take the logarithm of both sides of Eq~\ref{eq:constraint2} to obtain the final objective for one class label.
Together with Eq~\ref{eq:fisher_logic_multiplex_simplified}, we then obtain the final logical constraint in the main text in Eq~\ref{eq:cifar100_domain_knowledge}.

\begin{equation}
    \label{eq:constraint3}
     y_j > \log(\frac{1-\epsilon}{\epsilon}) + \log \sum\limits_{i \notin SC_{people}} e^{y_i}
\end{equation}

This implementation can then be directly encoded into the MultiplexNet loss as usual (where $y_k$ refer to the constrained output of the network for each of the $20$ super classes, $\pi_k$ is the MultiplexNet probability for selecting logic term $k$, and $CE$ is the standard cross entropy loss that is used in image classification.

\begin{equation}
    \label{eq:image_loss}
     \mathcal{L}(\theta) = \sum\limits_{i=1}^{20} \pi_k \left[ CE(y_k) + \log \pi_k \right]
\end{equation}

\textbf{Network Architecture}

We use a Wide ResNet 28-10~\citep{zagoruyko2016wide} in all of the experimental conditions for this CIFAR100 experiment.
We build on top of the pytorch implementation of the Wide ResNet that is available here: \url{https://github.com/xternalz/WideResNet-pytorch}.
This implementation is available under an MIT License.

\section{Appendix B: MultiplexNet Architecture Overview}
\label{appendix:architecture_overview}

MultiplexNet accepts as input a data set consisting of samples from some target distribution, $p^*(x)$, and some constraints, $\Phi$ that are known about the data set. 
We assume that the constraints are correct, in that Eq.~\ref{eq:domain_constraints} (main text) holds for all $x$.
We aim to model the unknown density, $p^*$, by maximising the likelihood of a parameterised model, $p_\theta(x)$ on the given data set.
Moreover, our goal is to incorporate the domain constraints, $\Phi$, into the training of this model.

We first assume that the domain constraints are provided in DNF.
This is a reasonable assumption as any logical formula can be compiled to DNF, although there might be an exponential number of terms in worst case scenarios (as discussed in Section: Limitations).
For each term $\phi_k$ in the DNF representation of $\Phi = \phi_1 \lor \phi_2 \lor \dots \lor \phi_K$, we then introduce a transformation, $h_k$, that ensures any real-valued input is transformed to satisfy that term.
Given a Softplus transformation $g$, we can suitably restrict the domain of any real-valued variable such that the output satisfies $\phi_k$.
For example, consider the constraint, e.g., $\phi_1: x > y + 2 \land x < 5$.
The transformation $h_1(x') = -g(-(g(x') + \alpha) + \beta)$ will constrain the real-valued variable $x'$ such that $\phi_1$ is satisfied. 
In this example, $y$ does not need to be constrained.
Here $\beta = 5$ and $\alpha = \log( e^{5 - (y + 2)} - 1)$.
Any combination of inequalities can be suitably restricted in this way.
Equality constraints, can be handled by simply setting the output to the value that is specified.

MultiplexNet therefore accepts the unconstrained output of a network, $x' \in \mathbb{R}$, and introduces $K$ constraint terms $h_k$ that each guarantee the constrained output $x_k = h(x')$ will satisfy a term, $\phi_k$, in the DNF representation of the constraints.
The output of the network is then $K$ transformed versions of $x'$ where each output $x_k$ is guaranteed to satisfy $\Phi$.
The Categorical selection variable $k \sim q(k \mid x)$ can be marginalised out leading to the following objective:

\begin{equation}
    \label{eq:multiplexnet_objective}
     \mathcal{L}(\theta) = \sum\limits_{i=1}^{20} \pi_k \left[ \mathcal{L'}(x_k) + \log \pi_k \right]
\end{equation}

\begin{figure*}
  \centering
  \includegraphics[width=\textwidth]{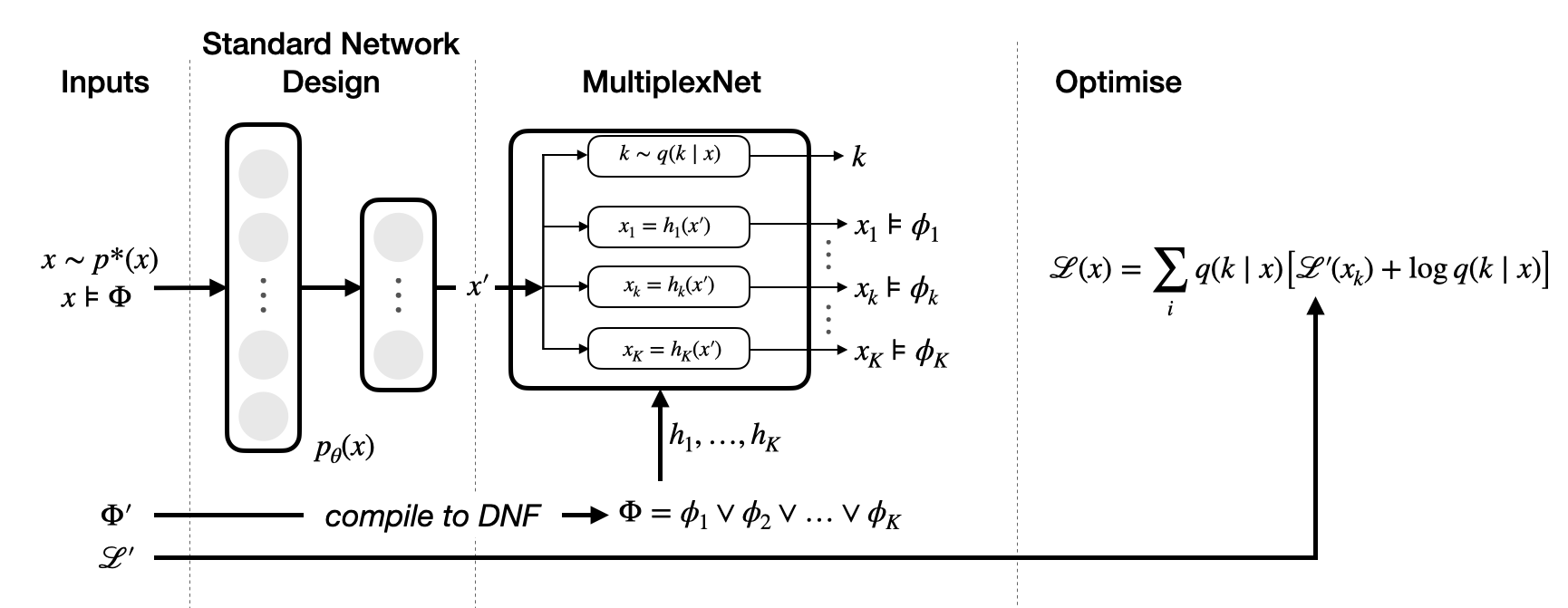}
  \caption{Architecture of the MultiplexNet. We show how to append this framework to an existing learning scheme. The unconstrained output of the network $x'$, along with the constrain transformation terms $h_1, \dots , h_K$ are used to create $K$ constrained output terms $x_1, \dots , x_K$. The latent Categorical variable $k$ is used to select which term is active for a given input. In this paper, we marginalise the Categorical variable leading to the specified loss function.}
  \label{fig:pictorial_architecture}
\end{figure*}

In Eq.~\ref{eq:multiplexnet_objective}, $\mathcal{L'}$ refers to the observation likelihood that would be used in the absence of any constraint. 
$x_k$ is the $k^{th}$ constrained term of the unconstrained output of the network: $x_k = h_k(x')$.
This architecture is represented pictorially in Fig.~\ref{fig:pictorial_architecture}.

\end{document}